%% file: main.tex
\documentclass[letterpaper, 10 pt, conference]{ieeeconf}  

\IEEEoverridecommandlockouts                              

\input{preamble.tex}

\usepackage{caption}
\newcommand{\citep}[1]{\cite{#1}}
\usepackage{authblk}

\title{\LARGE \bf
Learning for Microrobot Exploration: Model-based Locomotion, Sparse-robust Navigation, and Low-power Deep Classification
}

\author{Nathan O. Lambert\textsuperscript{1}, Farhan Toddywala\textsuperscript{1}, Brian Liao\textsuperscript{1}, Eric Zhu\textsuperscript{1}, Lydia Lee\textsuperscript{1}, and Kristofer S. J. Pister\textsuperscript{1}

\thanks{Corresponding author: Nathan O. Lambert, \tt \href{mailto:nol@berkeley.edu}{nol@berkeley.edu}}%
\thanks{\textsuperscript{1}Department of Electrical Engineering and Computer Sciences, University of California, Berkeley.}
}

\begin{document}


\maketitle



\begin{abstract}
\input{0_abstract.tex}
\end{abstract}




\input{1_introduction.tex}

\input{2_relatedwork.tex}

\input{3_low_level_control.tex}

\input{5_efficient_slam.tex}
\input{6_low_power_classification}

\input{8_conclusion.tex}


\section*{ACKNOWLEDGMENT}
The authors thank the Berkeley Sensor \& Actuator Center.

\bibliographystyle{IEEEtran}
\bibliography{main}

\input{_appendices.tex}

\end{document}

%% file: preamble.tex
\usepackage{graphicx} 
\usepackage{amsmath} 

\usepackage{csvsimple}  

\usepackage{tikz}
\usetikzlibrary{positioning, shapes, arrows.meta}
\usetikzlibrary{shapes,arrows,chains}
\usetikzlibrary{arrows,calc,automata}

\usepackage{subcaption}

\usepackage[ruled,vlined]{algorithm2e}
\usepackage{algorithmic}

\usepackage{siunitx}

\usepackage{afterpage}
\usepackage{placeins}
\usepackage{todonotes}

\usepackage{verbatim}
\usepackage{hyperref}
\usepackage{nameref}
\usepackage{textcomp}
\usepackage{gensymb} 
\usepackage{siunitx}

\tikzset{
    block/.style = {draw, rectangle, 
        minimum height=.5cm, 
        minimum width=1cm},
    input/.style = {coordinate,node distance=.5cm},
    output/.style = {coordinate,node distance=.5cm},
    arrow/.style={draw, -latex,node distance=.5cm},
    pinstyle/.style = {pin edge={latex-, black,node distance=1cm}},
    sum/.style = {draw, circle, node distance=1cm},
    line/.style={-{Stealth}}
    }

\usepackage{cite}

\usepackage{amsmath,amssymb}

\usepackage{array}

\usepackage{upgreek}
\usepackage{float}          
\usepackage{stfloats}

\usepackage{float}
\floatstyle{plain}
\restylefloat{table}
\usepackage{graphicx}

\newcommand{\fig}[1]{Figure~\ref{#1}}

\newcommand{\tab}[1]{Table~\ref{#1}}
\newcommand{\alg}[1]{Algorithm~\ref{#1}}
\usepackage{dsfont}
\usepackage{amsmath}

\newcommand{\prob}{{p}} 					

\newcommand{\D}[0]{\mathcal{D}} 				



%% file: 0_abstract.tex
Building intelligent autonomous systems at any scale is challenging.
The sensing and computation constraints of a microrobot platform make the problems harder.
We present improvements to learning-based methods for on-board learning of locomotion, classification, and navigation of microrobots. 
We show how simulated locomotion can be achieved with model-based reinforcement learning via on-board sensor data distilled into control. 
Next, we introduce a sparse, linear detector and a Dynamic Thresholding method to FAST Visual Odometry for improved navigation in the noisy regime of mm scale imagery. 
We end with a new image classifier capable of classification with fewer than one million multiply-and-accumulate (MAC) operations by combining fast downsampling, efficient layer structures and hard activation functions. 
These are promising steps toward using state-of-the-art algorithms in the power-limited world of edge-intelligence and microrobots.

%% file: 1_introduction.tex
\section{Introduction}
Microrobots have been touted as a coming revolution for many tasks, such as search and rescue, agriculture, or distributed sensing \cite{flynn1987gnat,brambilla2013swarm}.
Microrobotics is a synthesis of Microelectromechanical systems (MEMs), actuators, power electronics, and computation. 
Recent work has demonstrated progress towards controlled walking and flying robots, with primary development on the actuators \cite{drew2018toward, contreras2017first, jafferis2019untethered}. 

The above applications require the micro-agents to explore the world without supervision.
Exploration entails deciding where to go, how to move, encoding where the robot has been, and retaining  information about the world.
Recent results in controlled flight, first steps, miniaturized radios communicating, and at-scale power electronics have set up the next step: intelligent micro-agents.
The capabilities of each sub-component are growing, but little has been done to integrate these into autonomous systems. 
Applying breakthroughs in intelligence on microrobots requires focusing on different problems than peak accuracy or higher dimensional modelling: microrobots have little memory to store data, limited compute and battery power for computation, and suffer from noisy sensors.
In this paper, we showcase improvements to low-power machine learning techniques that could scale to microrobot exploration. 

\begin{figure}[t]
    \centering
    \includegraphics[width=\columnwidth]{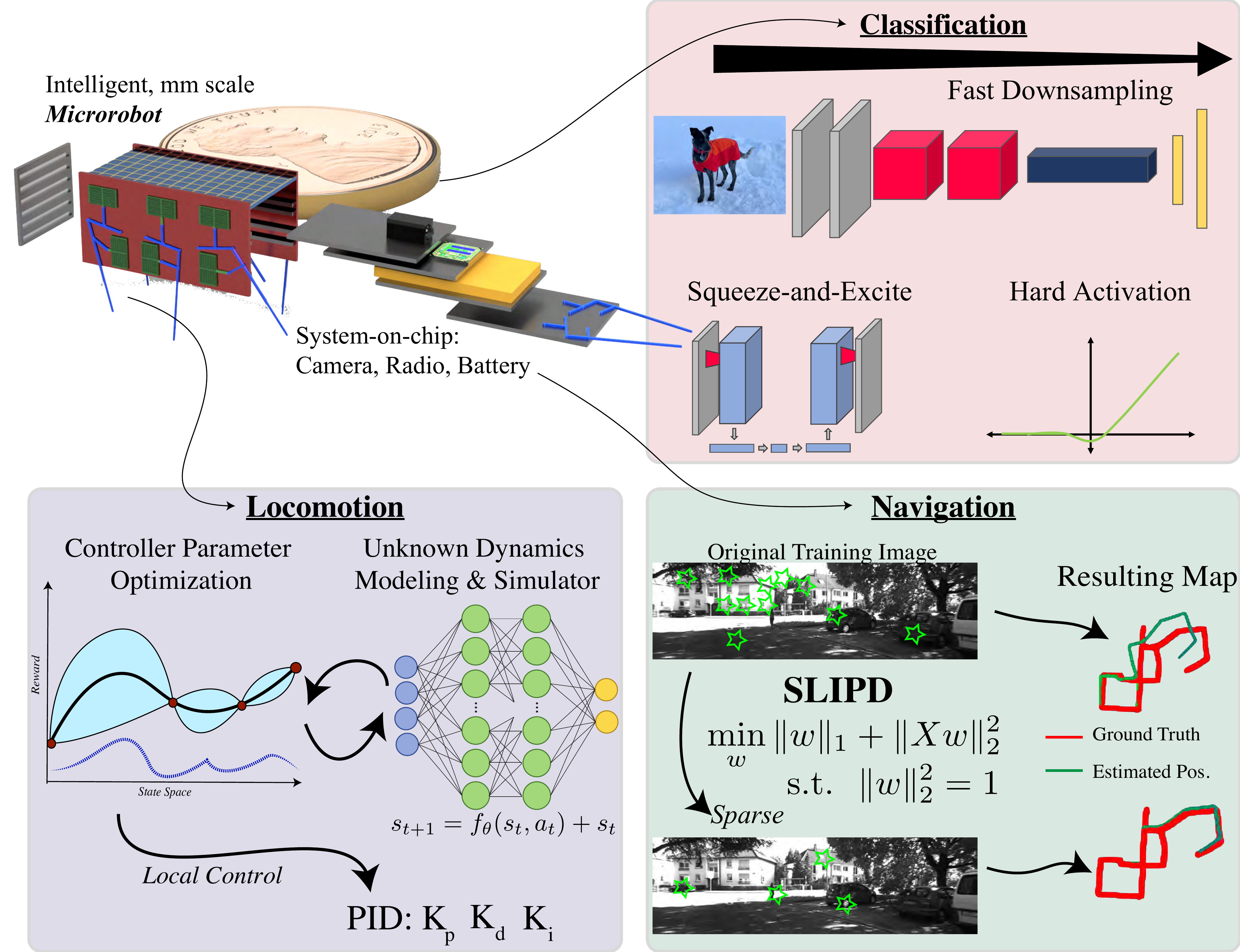}
    \caption{Our vision for microrobot exploration based on three contributions: 
    1) improving data-efficiency of learning control, 
    2) a more noise-robust and novel approach to visual SLAM (SLIPD), and 
    3) state-of-the-art deep learning based classifier at power budgets for microrobots.}
    \label{fig:vision}
    \vspace{-4mm}
\end{figure}

The pillars of autonomy for micro-robots in this paper are highlighted in \fig{fig:vision}, being controller generation for \textbf{\textit{locomotion}}, visual Simultaneous Localization and Mapping (SLAM) for \textit{\textbf{navigation}}, and compressed deep learning for \textbf{\textit{classification}}.
First, we present how sample-efficient model-based reinforcement learning can generate generalizable locomotion primitives in data-constrained applications.
Next, with the ability to explore, we detail a robust visual-SLAM approach applying online hyperparameter tuning and unsupervised learning to keypoint detection in order to enable visual-SLAM on-board.
We conclude with demonstrating the ability for a single chip to classify basic objects on chip -- we coin the new network with under 1 million multiply-and-accumulate (MAC) operations MicroBotNet. 

%% file: 2_relatedwork.tex
\section{Related Work}

\subsection{Microrobots}
Microrobots have been taking off and taking steps in recent years. 
Walking robots are capable of a higher payload and longer movement time, due to their low speed and force requirements. 
Multiple walkers have emerged with bio-inspired designs mirroring insects \cite{zhang2011bioinspired, contreras2017first}. 
Similarly, there are biologically inspired, flapping wing fliers \cite{ zou2016liftoff,  jafferis2019untethered}.
Recently other mechanisms, such as electrohydrodynamic force, have been shown to operate at the force to mass ratios required for sustained flight \cite{drew2018toward}.
Unlike the passively-stable walkers, the flying robots represent more challenging control problems.

\begin{figure}[t]
    \centering
    \begin{subfigure}{0.32\columnwidth}
        \centering
        \includegraphics[width=\linewidth]{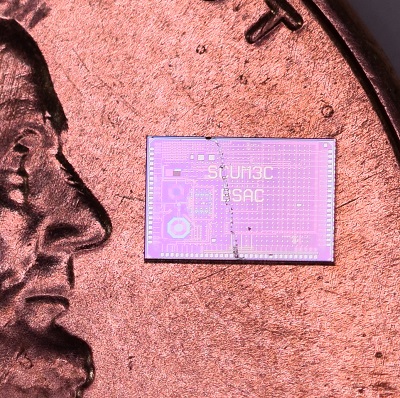}
        \caption{{\small SC\si{\micro}M \cite{scum}}}    
        \label{fig:scum}
    \end{subfigure}
    \hfill
    \begin{subfigure}{0.32\columnwidth}  
        \centering 
        \includegraphics[width=\linewidth]{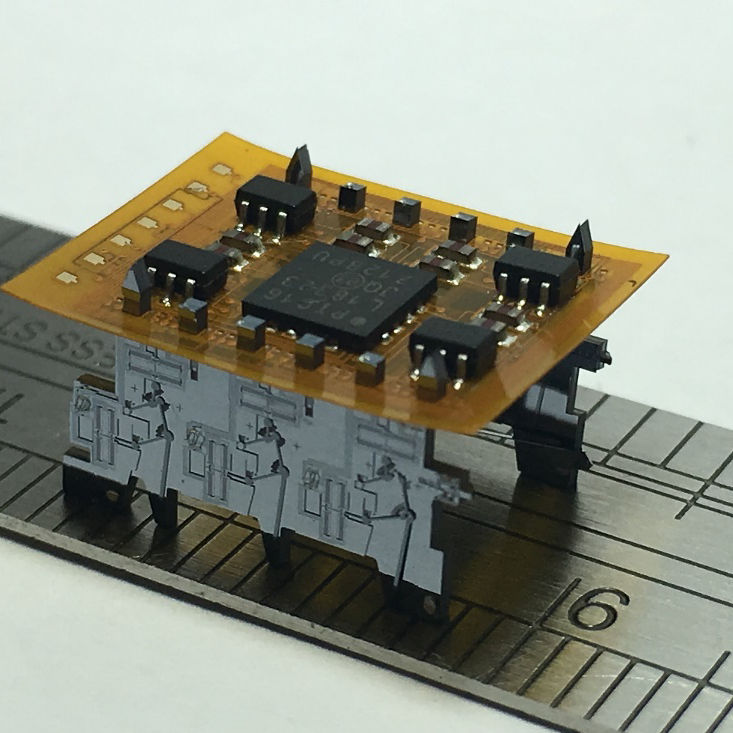}
        \caption{{\small Silicon Walker \cite{contreras2017first}}}    
        \label{fig:walker}
    \end{subfigure}
    \hfill
    \begin{subfigure}{0.32\columnwidth}   
        \centering 
        \includegraphics[width=\linewidth]{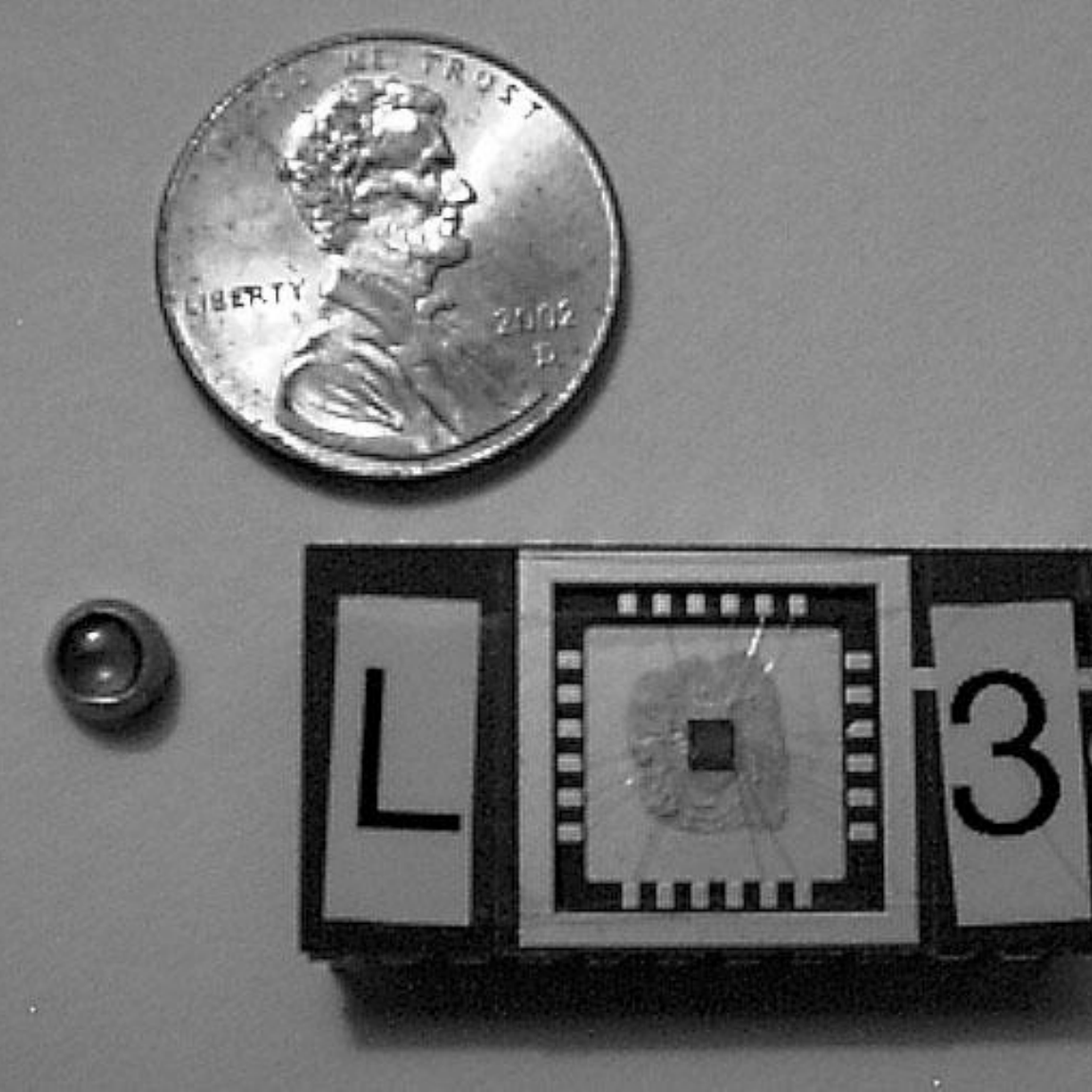}
        \caption{{\small  \SI{}{\milli \meter^3} imager\cite{Choy2003Camera}}    }
        \label{fig:mockup}
    \end{subfigure}
    \vspace{-1mm}
    \caption{Pieces of the micro-robot platform. 
    Left) a single-chip mote with integrated radio TRX, microprocessor, and sensor interface; center) a silicon walking microrobot, and right) an initial $2$\si{\milli \meter}$^3$ camera capable of $128\text{x}128$ pixel images. 
    }

    \label{fig:robotss}
    \vspace{-4mm}
\end{figure}

Building out the platform for autonomous microrobots has been a series of system-on-chip (SoC) breakthroughs towards \SI{}{\milli\meter} scale computation, sensing and power. 
Chips on the scale of \SI{1}{\milli \meter}$^2$ have been created for \SI{100}{V}, multi-channel power supply \cite{Rentmeister_2020} and mesh-networked radio communication \cite{scum}.
A \SI{2}{\milli \meter^3} camera can be affixed to modern microrobots and take $128 \times 128$ greyscale images \cite{Choy2003Camera, hanson20100}.
Some of the robots and components 
that motivate
this investigation are shown in \fig{fig:robotss}.
Integrating these breakthroughs represents the hurdle to creating autonomous microrobots. 

\subsection{Low-level Controller Synthesis} 
Designing low-level controllers for unknown systems poses a challenge -- a balance between data efficiency, safety, and effectiveness. 
Model-based reinforcement learning (MBRL) has emerged as a useful candidate in multiple small-scale robotic platforms: including quadrupeds \cite{nagabandi2017neural}, micro-aerial-vehicles \cite{lowlevelmbrl}, and remote-control cars \cite{williams2017information}.
MBRL methods offer encouragement over standard approaches such as system identification (SI) or proven state-space controllers (LQR, iLQR) because of MBRL's ability to capture unmodeled dynamics, such as tethers or process variation.
MBRL works in an iterative process of gathering data about the environment, forming a model $p_\theta(s_t,a_t)$ with the data, and leveraging said model to learn a controller.

While state-of-the-art MBRL algorithms showcase strong asymptotic performance \cite{janner2019trust, chua2018deep}, the computational requirements include substantial data storage to form the model and low-frequency control -- even on a graphics processing unit. 
Changes to the data storage and control policies need to be made for these devices to run on edge-devices.
Improved methods of controller synthesis are needed for microrobots because they 1) lack analytical dynamics models offline control design and 2) have a high cost per test, yielding motivation for  prior-free and safe methods.

\subsection{Robust SLAM}
Simultaneous localization and mapping (SLAM) techniques have emerged as the algorithmic foundation for creating a reference map as agents explore and interact with the world~\cite{bailey2006simultaneous, durrant2006simultaneous}.
Most SLAM algorithms work by tracking a set of keypoints with linear state estimators (\textit{e.g.} Kalman filters) as they move through space to create a grounded, feature-based and local map.
With the growth of computer vision, visual odometry -- algorithms with primary SLAM data derived from images -- has emerged as a popular candidate. 
A common solution, Features from Accelerated Segments Test (FAST)~\cite{rosten2005fast} is computationally efficient and simple to implementation.
Visual odometry returns a pose estimate by tracking the movement of detected interest points across images ~\cite{scaramuzza2011visual}. 
Visual SLAM techniques degrade with noise~\cite{scaramuzza2011visual}, and have encountered difficultly translating to dynamic environments or low-quality cameras.
In such a realistic, noisy environment, we show that a Dynamic Thresholding method for computing interest points outperforms the vanilla FAST detector on SLAM tasks.

\subsection{Low Power Classification}
Deep Learning training and inference often incur significant computing and power expense, making them impractical for edge devices. 
Multiply-and-accumulates (MACs) are an accepted computational-cost metric as they map to both the multiply-and-accumulate computation and its memory access patterns of filter weights, layer input maps, and partial sums for layer output maps. 
Prior work has decreased parameter size and MAC operations.

SqueezeNet \cite{iandola2016squeezenet} introduced Fire modules as a compression method in an effort to reduce the number of parameters while maintaining accuracy. 
MobileNetV1 \cite{howard2017mobilenets} replaced standard convolution with depth-wise separable convolutions where a depth-wise convolution performs spatial filtering and pointwise convolutions generate features.
Fast Downsampling \cite{qin2018fd} expanded on MobileNet for extremely computationally constrained tasks--32$\times$ downsampling in the first 12 layers drops the computational substantially with a 5\% accuracy loss.
Trained Ternary Quantization \cite{zhu2016trained} reduced weight precision to 2-bit ternary values with scaling factors with zero accuracy loss.
MobileNetV3 \cite{howard2019searching} used neural architecture search optimizing for efficiency to design their model. 
Other improvements include `hard'~activation functions (h-swish and h-sigmoid) \cite{ramachandran2017searching}, inverted residuals and linear bottlenecks \cite{sandler2018mobilenetv2}, and squeeze-and-excite layers\cite{hu2018squeeze} that extract spatial and channel-wise information.

Benchmarking from a 45\si{\nano\meter} process \cite{horowitzenergy}, shrinking process nodes and decreased bit precision enable a MAC cost approaching 1\si{\pico\joule}. 
Targeting 1\si{\micro\joule} per forward-pass, we combine these advancements into a new network with $<$1 million MACs. 
Bankman, et. al \cite{bankman2018always} show a 3.8\si{\micro\joule} 86\% accuracy Cifar-10 classifier on-chip using a BinaryNet and a weight-stationary, data-parallel architecture with input reuse.
To estimate the energy cost of doing image processing remotely,  we can assume using a 1\si{\micro\watt} RF radio can transmit at 1 Mbps at 1 nJ/bit.
For a full 128 $\times$ 128 $\times$ 8 bit image multiplied by a factor of 10 for end-to-end networking, the total energy cost is $1$\si{\milli\joule}, significantly above $1$\si{\micro\joule}.

%% file: 3_low_level_control.tex
\section{Low-level control}
To begin, we detail the recent works in robot learning suited for microrobot tasks, and show how data efficiency -- and therefore power efficiency -- can be improved.
\subsection{Experimental Setting}
Low-level control of microrobots entails overcoming fabrication variation and testing risk.
Test setups of such robots include: difficult to model tethers, damaged mechanisms, changing environments (\textit{e.g.} different material surfaces).
Due to the variability of each robot and test environment, methods derived from a state-space formulation (\textit{e.g.} iLQR) are leveraging incomplete, and therefore sub-optimal, models.
The gain in modeling with our approach comes at the cost of any stability analysis through feedback control.

Deep neural networks are a proven candidate for modeling the nonlinear dynamics in a sample efficient manner \cite{lowlevelmbrl}.
Learning an environmental model as a distribution $s_{t+1} = p_\theta(s_t,a_t)$ can capture nonlinearities and unexpected variation with only $1000$ training points.
With a model, Bayesian Optimization can be used to optimize an underlying reward of a task, such as walking forward in~\cite{yang2018learning}.
Bayesian Optimization can solve for simple parametric controllers (\textit{e.g.} PID, sinusoidal patterns)~\cite{yang2018learning} in lieu of computationally intensive, sampling-based model-predictive control~\cite{lowlevelmbrl}.
Training models on experimental systems is a challenge in learning-based approaches because model accuracy suffers when there are anomalous, non-physical data (sensor aberrations) or skewed distributions.
State-of-the-art algorithms do not include steps to filter data in search of generalization and maximum sample efficiency \cite{chua2018deep, janner2019trust} -- but such a direct training process can result in numerical instability in prediction and low effectiveness in terms of maximizing reward \cite{mismatch2019}.
We propose intelligent data aggregation towards adaption on difficult physical problems.

The dataset presented is from a Crazyflie's~\cite{bitcraze2016crazyflie} on-board MPU9250 inertial measurement unit.
This platform suffers less than microrobots from sensor degradation, nonlinear dynamics, and test risk, so there is less potential for improvement and filtering.
With the smoother data, our experiments still show substantial improvement in the models over standard unfiltered training approaches.

\subsection{Improved Data Efficiency}
\begin{figure}[t]
    \centering
    \vspace{-7.5pt}
    \includegraphics[width=.95\columnwidth]{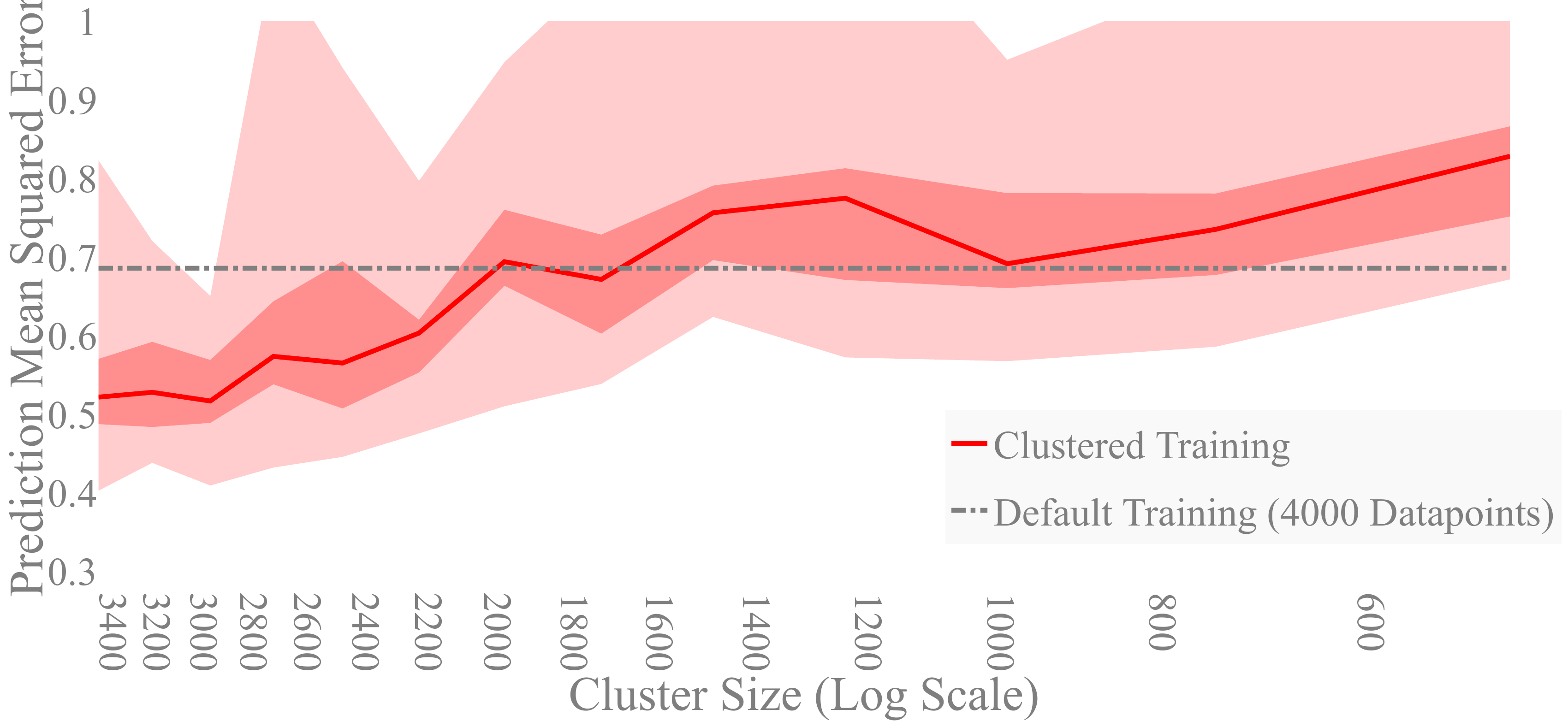}
    \vspace{-1.5mm}
    \caption{Removing redundant datapoints from quadrotor dynamics data can improve validation set accuracy while reducing stored data by over 50\%. 
    $M=25$ models were trained at each dataset size, and evaluated on the same validation set.
    From left to right is increasing the filtering, which improves the accuracy on an 800 point validation set by up to 25\%, but over-filtering begins to lose expressivity.
    }
    \label{fig:clustering}
    \vspace{-4mm}
\end{figure}

To alleviate the challenges of modelling experimental dynamics, we propose using clustering to filter out redundant data.
Current learning methods aggregate data from all trials, and because the tasks are repeated there is a large concentration of training data around initial states.
The training process focuses the model accuracy around the most frequent data.
\fig{fig:clustering} shows an improvement in validation set accuracy when training on a k-means clustered, uniformly representative subset of the training data.
This initial improvement can be scaled to other model types such as the Gaussian Processes, but more computationally intensive filtering methods should be considered.
For example, re-weighting training distributions by an expert trajectory improves sample efficiency in MBRL control tasks \cite{mismatch2019}.
A summary of the iterative model learning \& control building approach is shown in \alg{alg:mbrl}. 
\begin{algorithm}[h]
        \KwData{Initialize $\D$ from random actions}
        \While{Improving}{
            Train model $\prob_\theta(s_{t+1}|s_t, a_t)$ on $\D$\\
            Distill model $\prob_\theta(s_{t+1}|s_t, a_t)$ into controller $\pi(s_t)$ \\
            Collect data $\D'$ with $\pi(s_t)$ in real environment  \\
            Aggregate dataset $\D = \widehat{\D} \cup \D'$ \\ 
            Filter to optimal dataset $\widehat{\D} = f(\D)$
            \vspace{0.025in}
        }
    \caption{On-Device Model-based RL
    }
    \label{alg:mbrl}
\end{algorithm}

%% file: 5_efficient_slam.tex
\section{Robust SLAM}
Here we detail improvements towards the state-of-the-art visual odometry technique, FAST, and propose a new self-supervised learning method for SLAM.

\subsection{Experimental setup}
\begin{figure}[t]
    \centering
    \includegraphics[width=.85\columnwidth]{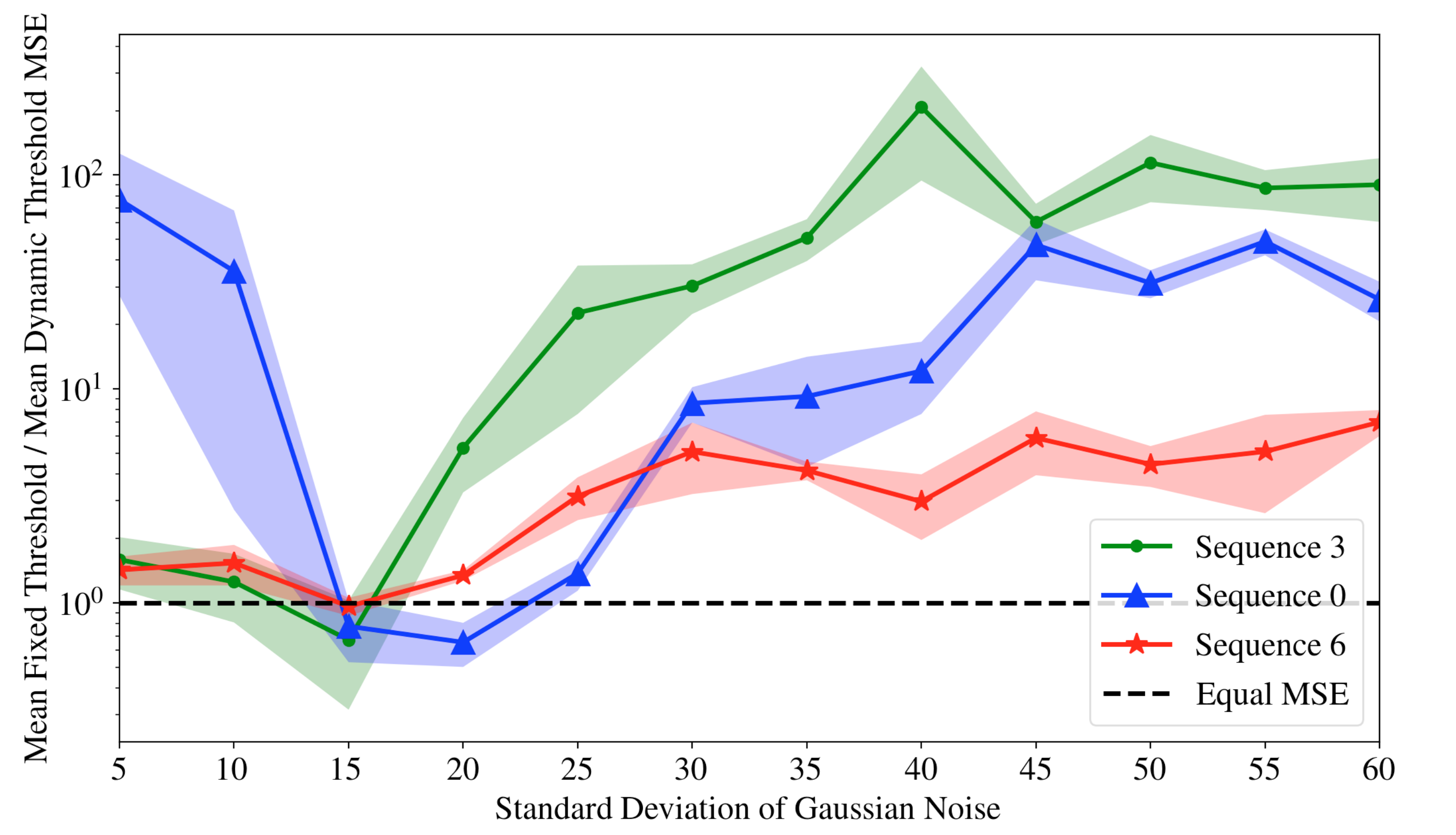}
    \caption{\textbf{Static Noise}: ratio of standard FAST error over average Dynamic Thresholding FAST error ($n=10$), higher ratio is lower error, with static i.i.d. Gaussian noise of intensities$\in[5,60]$ on KITTI sequences 0, 3, 6. 
    The lower noise levels are less consistent, but still show an improvement with Dynamic Thresholding  when the noise levels are constant.
    The dynamics thresholding shows improvement as the noise levels continue to increase beyond $\sigma = 25$.
    }
    \label{fig:nav1}
    \vspace{-4mm}
\end{figure}

\begin{figure}[t]
    \centering
    \includegraphics[width=.85\columnwidth]{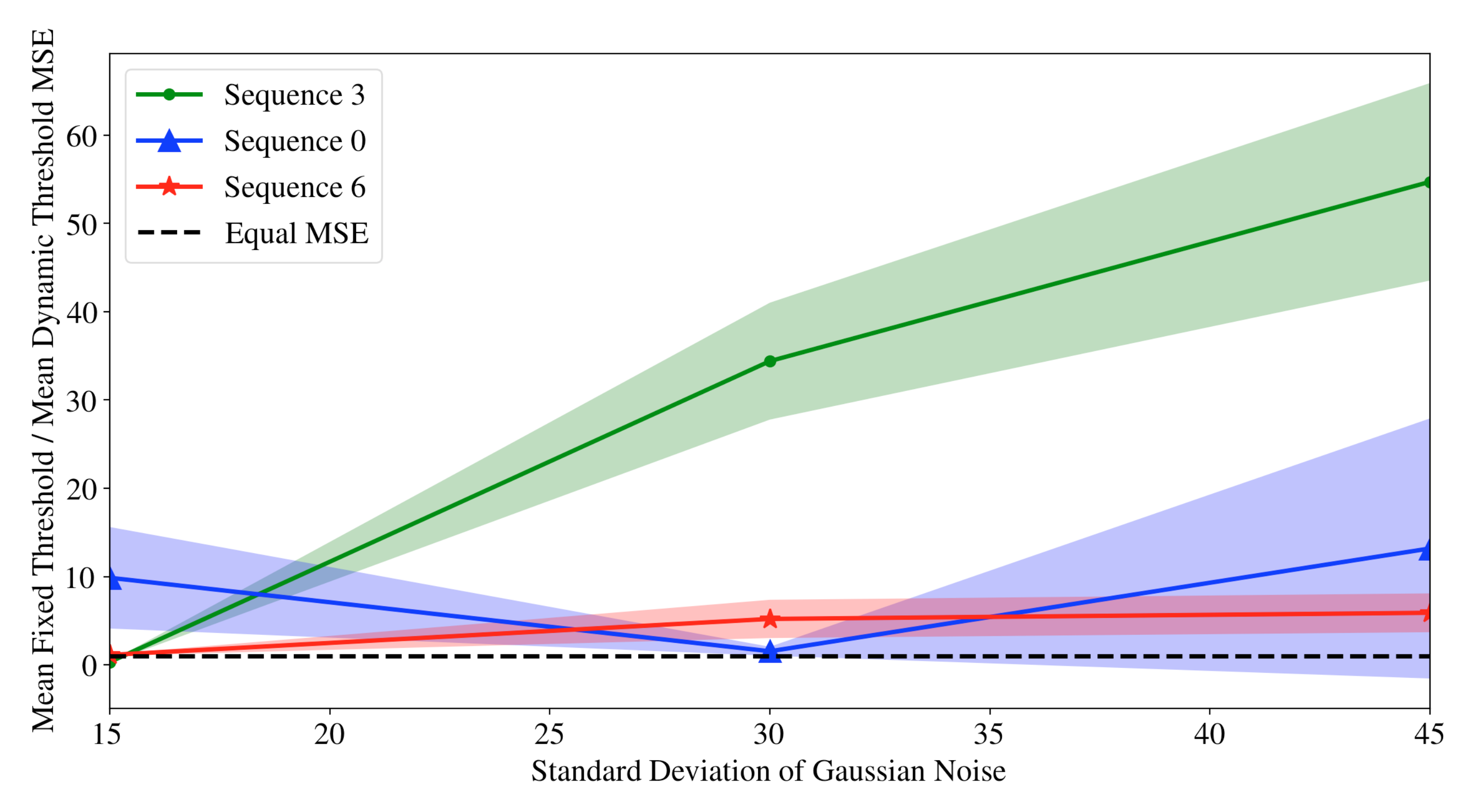}
    \caption{\textbf{Dynamic Noise}: ratio of standard FAST error over average Dynamic Thresholding error ($n=10$), higher is better, with dynamically changing additive noise over KITTI sequences 0, 3, 6. 
    The x-axis is the maximum noise level in the dynamic setting, $L$.
    There is a reduction in error of up to $50\times$ depending on the noise level and sequence.}
    \label{fig:nav2}
    \vspace{-4mm}
\end{figure}

\begin{figure}[t]
    \centering
    \begin{subfigure}{0.4\columnwidth}
        \centering
        \includegraphics[width=\linewidth]{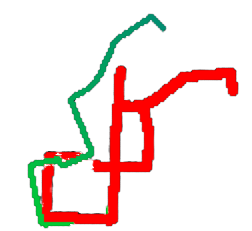}
        \caption{{\small Fixed Threshold}}    
        \label{fig:ex1}
    \end{subfigure}
    \quad
    \begin{subfigure}{0.4\columnwidth}  
        \centering 
        \includegraphics[width=\linewidth]{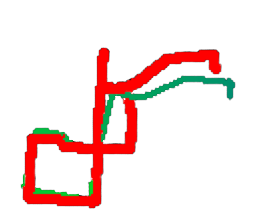}
        \caption{{\small Dynamic Thresholding}}    
        \label{fig:ex2}
    \end{subfigure}
    \hfill
    \vspace{-1.5mm}
    \caption{An example mapped trajectory from sequence 0 (about halfway through) with and without Dynamic Thresholding.
    The red trajectory in both images represents the ground truth trajectory and the green trajectory represents the estimated trajectory. 
    Left) standard FAST algorithm. 
    Right) FAST with Dynamic Thresholding.
    }

    \label{fig:Trajectory}
    \vspace{-4mm}
\end{figure}

To evaluate SLAM, we train on the KITTI Odometry Dataset of $1200 \times 375$ 8-bit greyscale images~\cite{Geiger2012CVPR}. 
We use sequences 3, 6 and 0 to represent easy, medium and hard trajectories. 
\fig{fig:Trajectory} shows a estimated versus true mapping on sequence 0.
We fix the FAST threshold at 50 for all experiments, chosen by cross-validating across accuracy with multiple thresholds. 
Our Dynamic Thresholding tunes this parameter online for improved performance, as shown in \fig{fig:nav1} and \fig{fig:nav2}.

Our experiments focus on the scenario of images with additional i.i.d. Gaussian noise across pixels.  
For our first set of experiments, we vary the standard deviation of the sampled additive Gaussian noise in $[0,60]$ pixels.
The noise levels used are higher than most \SI{}{\milli \meter} scale photography, but account for other process and computation errors~\cite{Choy2003Camera}.
We measure the Euclidean error between the predicted and the ground truth trajectory. 
Next, we model a sequence in which images are dynamically corrupted as the robot moves at different velocities--a varying noise level.
Similar to a random walk, we update the Gaussian noise intensity by adding a random pixel noise shift $X \in \{-1, 0, 1\}$ (discrete uniform distribution). 
The pixel noise is capped between 0 and some upper limit $L \in \{15, 30, 45\}$ to see how performance varies with different ranges of noise intensity during the sequence.

\subsection{Improved Keypoint Detection}

\paragraph{FAST Detector}
Features from Accelerated Segments Test (FAST) is a threshold based, corner detection algorithm. 
FAST is computationally efficient, but when noise is added the number of candidate corners spike erratically. 
Denoising methods are often computationally expensive and do not function at higher levels of noise.
Noise can emerge from electronic or other sources, and to transfer these approaches to the noisier sensors available on microrobots we develop more robust interest point detection.
\paragraph{Dynamic Thresholding}
To improve FAST, we propose a method to regulate the number of features without increasing the spatial or computational complexity of the algorithm.
Qualitatively, the automatic tuning sets a acceptable range of interest points. 
If FAST produces more than the upper limit of that range, we increase the threshold by a multiplicative constant greater than 1--increasing the FAST threshold will reduce the number of interest points. 
The threshold will be reduced similarly if FAST returns too few interest points.
This method is simple to implement and comes at very little cost in terms of processing and memory. 
The best values in our experiments are 1.1 and .9 for threshold updating rates, 1000 and 2000 for the range of acceptable interest point counts, and 50 for the base FAST threshold.
Dynamic Thresholding is sensitive to sharp movements, so the implementation could be improved to have an additional dynamic parameter acting in feedback from the magnitude of the pose-estimation update.

\subsection{Results}
Dynamic Thresholding shows a clear improvement over the standard FAST algorithm in the static noise (\fig{fig:nav1}) case when the standard deviation of the gaussian noise exceeds 20 and moderate improvements on dynamic, walking noise (\fig{fig:nav2}) with the upper limit of the gaussian noise exceeds 30. 
With static Gaussian noise with a standard deviation greater than 20, standard FAST trajectory MSE is on average $>$4.7 times higher for tested sequences than the Dynamic Thresholding trajectory. 
With dynamic Gaussian noise with an upper limit of 30 or greater, the standard FAST trajectory MSE is on average $>$5.5 times higher than the MSE with Dynamic Thresholding.
The results show that visual odometry can still be performed in noise heavy situations, which will be required to translate pose estimation to microrobots. 
The challenge with Dynamic Thresholding is the need for additional parameter tuning, so we propose an online, unsupervised approach  for full SLAM or as a companion to Dynamic Thresholding FAST.
A future comparison will be that between dynamic thresholding and a FAST method filtering the noisy raw pixel data over time.

\subsection{Unsupervised Keypoint Learning}
Unsupervised Learning for interest point detection has been proposed \cite{quadnet}, but none perform at a level where they are a viable replacement for FAST. 
Our approach formulates the detector as a sparse learning problem to minimize computation in determining keypoints. 
We compute a function $f_{\theta}$ (possibly parametrized by $\theta$ like an neural network) which produces the same score on 2 pixels in 2 images (after motion) representing the same point in 3D space. 
Then, we consider an $n \times n$ block around that pixel as candidate features for that pixel. 
Denote the set of features for point $i$ in 3D space in image 1 as $x_1^i$ and in image 2 as $x_2^i$. 
The optimization problem (solved via SGD) is then
$$\min_{\theta, w} \lambda||w||_1 + \frac{1}{n}\sum_{i = 1}^n (f_\theta(w \odot x_1^i) - f_\theta(w \odot x_2^i))^2$$
Additional constraints may be needed to ensure a nontrivial solution, such as tuning the sparsity weight via $\lambda$. 
Leaky Sparse Linear Interest Point Detector (Leaky-SLIPD), implements this method where $f$ applies a leaky-relu to the masked inputs and sums over them.
Preliminary results show moderate improvements over FAST on sequences 0 and 3 (though not on 6) when both algorithms use Dynamic Thresholding. See Appendix for details. $||w||_0 = 8$ in our experiments, resulting in an extremely efficient detector with fewer operations than FAST in the worst case.

%% file: 6_low_power_classification.tex
\section{Low Power Classification}
We discuss our new neural-network image classifier, MicroBotNet, capable of accurate prediction on low-resolution images with fewer than 1 million MAC operations

\begin{figure}[t]
    \centering
    \includegraphics[width=.8\columnwidth]{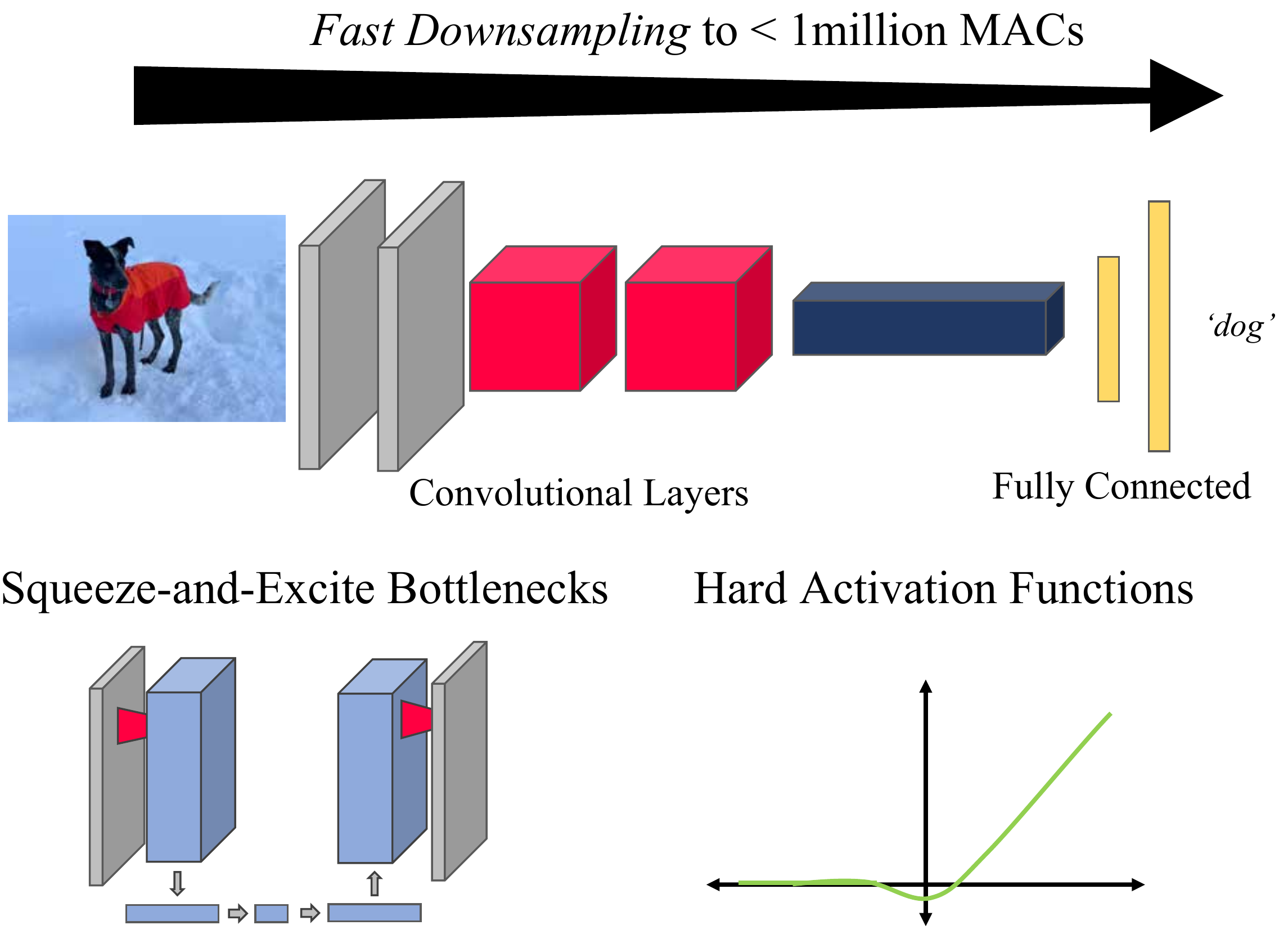}
    \caption{We introduce convolutional layers, for image processing, with larger kernel sizes early on to achieve fast downsampling and decrease the kernel size, allowing us to decrease MAC count. 
    Similar to MobileNet V3, we utilize squeeze-and-excite bottlenecks and hard activation functions for non-linear transformations.}
    \label{fig:cv}
    \vspace{-4mm}
\end{figure}

\begin{figure}[t]
    \centering
    \includegraphics[width=.85\columnwidth]{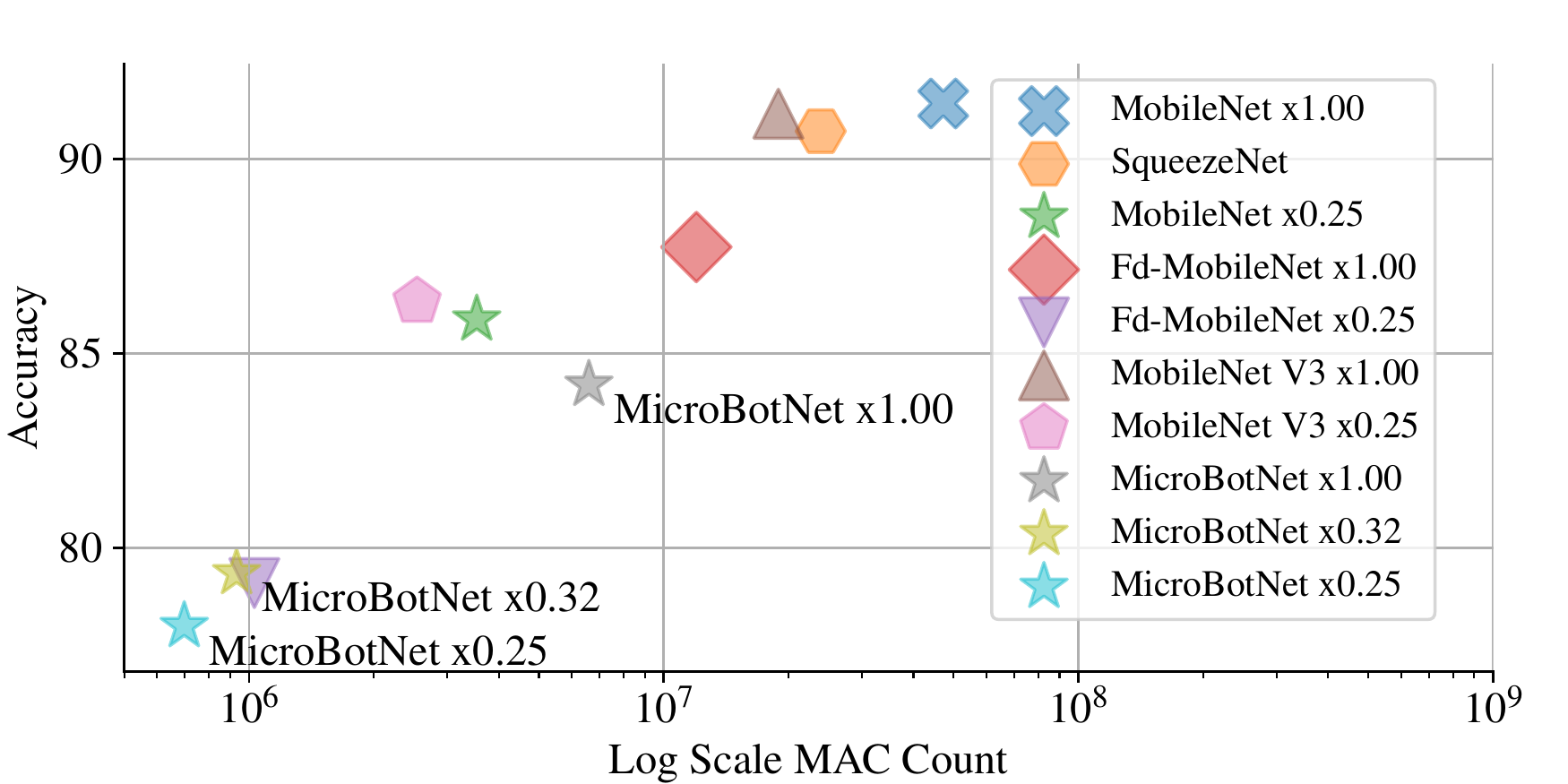}
    \caption{Tradeoff of MAC Count and Accuracy of different models we evaluated on CIFAR-10. 
    MicroBotNet $\times$0.32 improves on FD-MobileNet $\times$0.25 while continuing the trend of accuracy towards low-MACs.}
    \label{fig:classify}
    \vspace{-4mm}
\end{figure}

\subsection{Experimental Setup}

We evaluate the accuracy of energy-efficient neural networks on the CIFAR-10 dataset \cite{krizhevsky2014cifar} using an 12GB Nvidia K80 GPU.
The number of parameters and MAC operations are calculated with THOP \cite{thop}.
We use standard pre-processing including random cropping, random horizontal flipping, and normalization of training and testing images. 
For each model, we use stochastic gradient descent with $0.9$ momentum, $256$ batch size, $5\text{e-}4$ weight decay, and a learning schedule with a learning rate of $0.1$ and a decay of $0.1$ every $50$ epochs for $200$ epochs. 
We benchmark against SqueezeNet, MobileNet, FD-MobileNet, and MobileNet V3 on width multipliers of $\times$0.25 and $\times$1.00. 
The width multiplier increases or decreases the width of each layer, allowing for adaptation to the correct model for one's computational needs. 
We validate against our model, \textbf{MicroBotNet}, on width multipliers of $\times$0.25, $\times$1.00, and $\times$0.32.

\begin{table}[t]
    \centering
    \small{
    \begin{tabular}{|c|c|c|c|}
        \hline
        Model & Top-1 & MAC & Parameters\\\hline
        MobileNet V3 $\times$0.25 & 86.33 & 2,540,820 & 124,050\\
        MobileNet $\times$0.25 & 85.87 & 3,539,456 & 215,642\\
        \textbf{MicroBotNet $\times$0.32} & 79.35 &	932,886 & 236,658\\
        Fd-MobileNet $\times$0.25 & 79.09 & 1,029,888 & \textbf{128,730}\\
        \textbf{MicroBotNet $\times$0.25} & 77.99 & \textbf{697,662}	& 160,162\\\hline
    \end{tabular}}
    \caption{
    Top-1 is accuracy on CIFAR-10. 
    Comparison of MicroBotNet $\times$0.32 and $\times$0.25 with similar architectures.}
    \label{tab:comparisonMAC1}
\end{table}

\begin{table}[t]
    \vspace{-3mm}
    \centering
    \small{
    \begin{tabular}{|c|c|c|c|}
        \hline
        Model & Top-1 & MAC & Parameters\\\hline
        MobileNet $\times$1.00 & 91.42 & 47,187,968 & 3,217,226\\
        MobileNet V3 $\times$1.00 & 91.16 & 18,891,842 & 1,518,594\\
        SqueezeNet & 90.71 & 23,902,388 & \textbf{730,314}\\
        Fd-MobileNet $\times$1.00 & 87.73 & 11,983,872 & 1,886,538\\
        \textbf{MicroBotNet $\times$1.00} & 84.19 & \textbf{6,597,218} & 2,044,298\\\hline
    \end{tabular}}
    \caption{Top-1 is accuracy on CIFAR-10. 
    Comparison of MicroBotNet $\times$1.00 with similar architectures.
    \vspace{-6mm}}
    \label{tab:comparisonMAC2}
\end{table}

\subsection{MicroBotNet}

MicroBotNet applies Fast-Downsampling to MobileNetV3, summarized in \tab{tab:comparisonMAC1}.
MicroBotNet has 8$\times$ fast-downsampling in the first 6 layers 
because high dimension layers are the majority of forward pass computation cost. 
We include the width multiplier $\alpha$ which allows the model to generalize based on one's MAC computation needs. 
Width multipliers of $\times$0.25, $\times$1.00, and $\times$0.32 are included as reference. 
A minimum feature dimension of $4\times 4$ is set during downsampling to maintain suitable information capacity in our network, differentiating our down-sampling protocol from what is done on larger network designs. 
MicroBotNet leverages other low-power techniques, including squeeze-and-excite layers, h-swish and h-sigmoid, and inverted-residual and linear-bottlenecks layers.
We show a new downsampling schedule of bottleneck layers to meet microrobot computing capacity goals.

\subsection{Results}

MicroBotNet $\times$0.32 achieves 79.35\% accuracy while only using 932,886 MACs. 
This outperforms the previous work by 0.26\% in FD-MobileNet $\times$0.25, which achieves 79.09\% accuracy with 1,029,888 MACs. 
We also test MicroBotNet $\times$0.25 to compare to other standard fast-downsampling approaches, which achieves 77.99\% with 697,662 MACs. 
This work represents a further step in the trend of low-power classification shown in \fig{fig:classify}.

%% file: 8_conclusion.tex
\section{Conclusions}

This paper presents multiple learning-based steps towards integrated, autonomous microrobots. 
We outline the most data efficient methods for learning simple locomotion primitives in difficult to model regimes, navigation with noisy images, and classification with a 1\si{\micro\joule} forward-pass energy-cost.
The learning techniques all focus on data-efficiency and generalization to unknown, dynamic environments.
Further results, code, and video can be found on the website\footnote{ \href{https://sites.google.com/berkeley.edu/micro-autonomy}{https://sites.google.com/berkeley.edu/micro$\text{-}$autonomy}}.

In our vision of an autonomous future, microrobots play a critical role as edge-intelligent devices. 
This paper shows how state-of-the-art machine learning can be scaled to current on-chip capabilities.
All of these advances are conditioned on continued progress in research-grade batteries~\cite{ostfeld2016high}.
Each sub-problem poses a separate integration problem, but recent hardware and algorithmic research are capable of numerous currently untouched, impactful tasks.

%% file: _appendices.tex
\clearpage
\newpage

\section{Appendix}

\subsection{Clustering effect on distribution}
Here we show the effect of clustering on the dataset distribution of micro-quadrotor dynamics. 
Even a simple clustering algorithm k-means can work on low dataset sizes.
The clustering mechanism effectively flattens out the density of points in the distribution, resulting in a more uniform training environment for the neural network model.
\begin{figure}[h]
    \centering
    \begin{subfigure}{0.49\columnwidth}
        \centering
        \includegraphics[width=\linewidth]{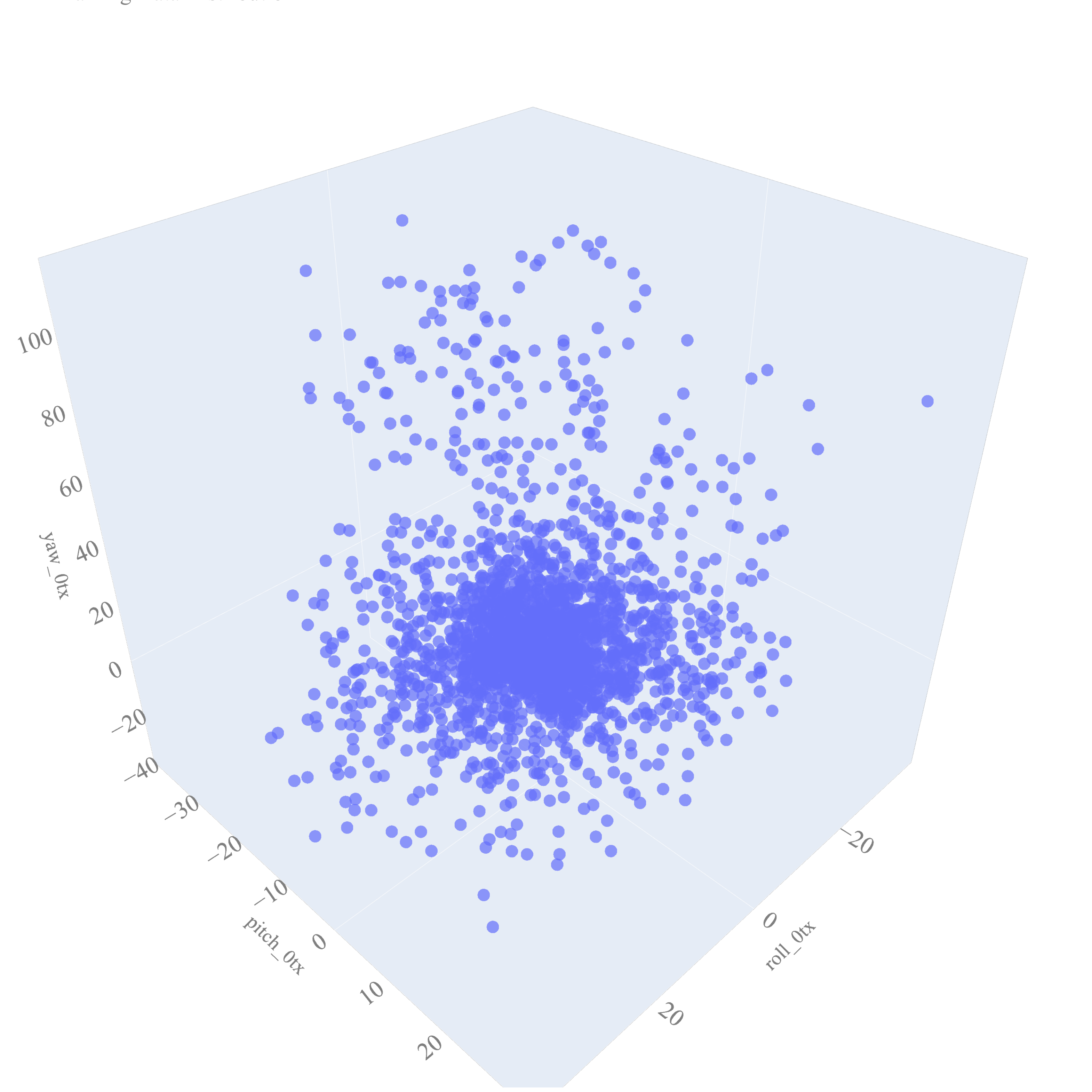}
        \caption{\small Original distribution \\ (4000 points).}    
        \label{fig:dist_full}
    \end{subfigure}
    \hfill
    \begin{subfigure}{0.49\columnwidth}  
        \centering 
        \includegraphics[width=\linewidth]{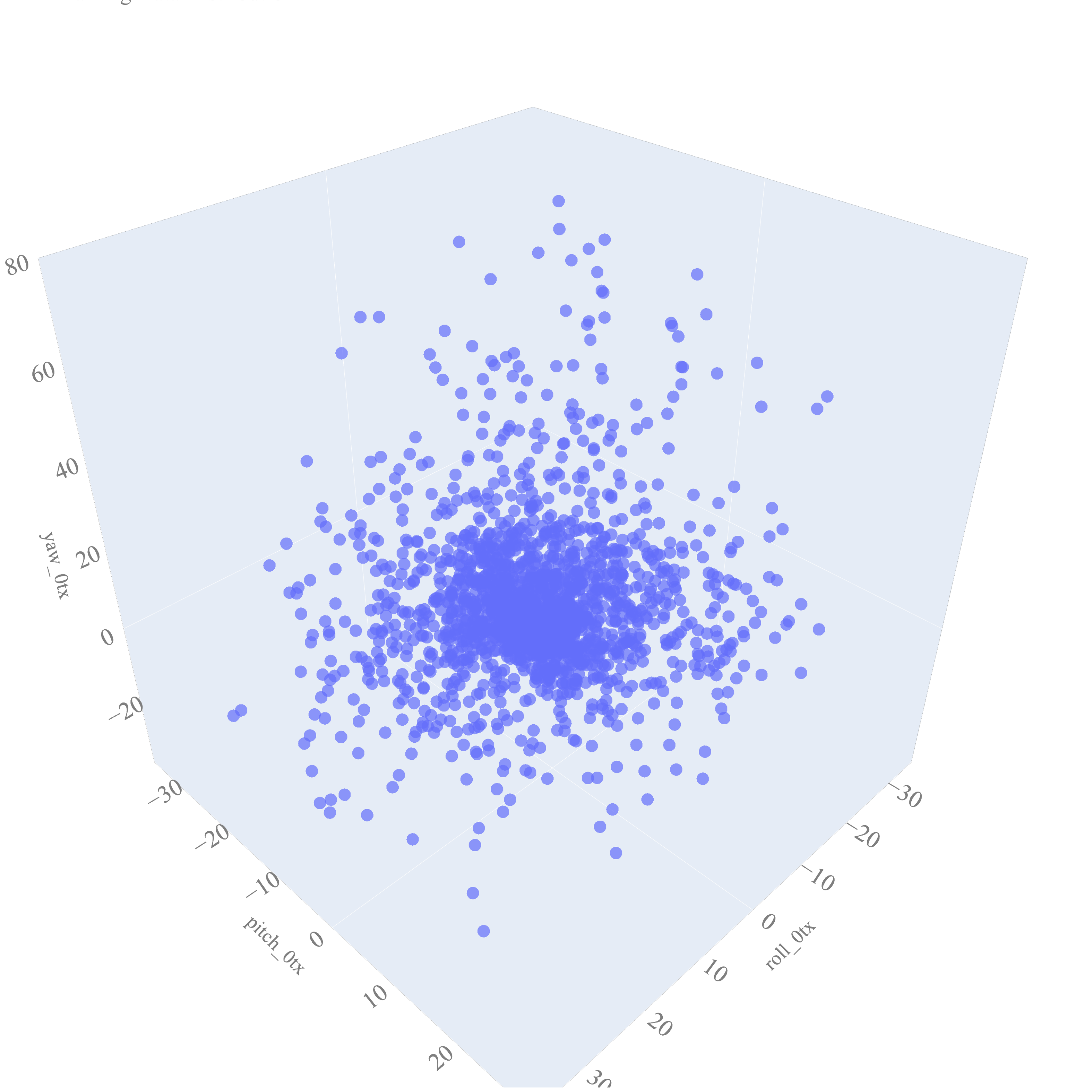}
        \caption{\small Clustered distribution \\ (500 points)}    
        \label{fig:dist_clust}
    \end{subfigure}
    \vspace{-1mm}
    \caption{Effect of clustering on training dataset. 
    Clustering can drastically reduce the number of data-points needed for training while maintaining, or even improving, validation set accuracy.
    The new distribution has approximately the same coverage, but with a more uniform density.
    }
    \label{fig:dist}
\end{figure}



\subsection{Hyperparameter Tuning}

\subsubsection{\textbf{Robust Keypoint Tuning}}
In practice, Dynamic Thresholding is primarily dependent on a good choice of a range of interest points. 
Some heuristics we can provide are that 1000 to 2000 worked well across many different sequences. 
Additionally, more gross levels of corruption beyond Gaussian noise with a standard deviation of 60 may require this range to be shifted up slightly to somewhere around 1500-2500. 
A relatively wide range of acceptable interest point counts allows for the threshold to remain stable; changing an already good threshold too often can result in poor performance.
The rates of change for the FAST threshold in Dynamic Thresholding are fairly robust given a good range of interest points. 
1.1 and 0.9 worked well in the experiments we performed, but not much difference was noticed between 1.25 and 0.8. 
These derivatives of dynamic threshold must pair nicely with the rate of change in the camera noise level.

\subsubsection{\textbf{MicroBotNet}}

When choosing hyperparameters, we experimented with different optimizers and training schedules.
We experimented with RMSprop with a lower learning rate of $.005$ that decays by $0.5$ every $50$ epochs and no momentum.
This did not converge to as high of an accuracy as SGD with momentum.
Using a higher learning rate and momentum on RMSprop would cause our gradients to explode and fail to converge.
For SGD and momentum, we used a training schedule with a learning rate of $0.1$ that decayed by $0.1$ every $50$ epochs.
This achieved a good balance between having a high learning rate and being able to have the learning rate decrease when loss converged.
Using recent advancements in automated parameter tuning, it is likely the accuracy and energy cost of our models could be improved further.

\subsection{MicroBotNet Model Specification}

\begin{table}[h]
    \centering
    \small{
    \begin{tabular}{|c|c|c|c|c|c|}
        \hline
        Input & Operator & exp size & \# out & SE & s\\\hline
        $32^2\times3$ & conv2d & - & 16 & - & 2\\
        $16^2\times16$ & bneck, $3\times3$ & 72 & 24 & No & 2\\
        $8^2\times24$ & bneck, $5\times5$ & 96 & 40 & Yes & 2\\
        $4^2\times40$ & bneck, $5\times5$ & 240 & 40 & Yes & 1\\
        $4^2\times40$ & bneck, $5\times5$ & 120 & 48 & Yes & 1\\
        $4^2\times48$ & bneck, $5\times5$ & 144 & 48 & Yes & 1\\
        $4^2\times96$ & bneck, $5\times5$ & 288 & 96 & Yes & 2\\
        $2^2\times96$ & bneck, $5\times5$ & 576 & 96 & Yes & 1\\
        $2^2\times96$ & bneck, $5\times5$ & 576 & 96 & Yes & 1\\
        $2^2\times96$ & bneck, $5\times5$ & 576 & 96 & Yes & 1\\
        $2^2\times96$ & bneck, $5\times5$ & 576 & 96 & Yes & 1\\
        $2^2\times96$ & conv2d, $1\times1$ & - & 576 & Yes & 1\\
        $2^2\times576$ & pool, $2\times2$ & - & - & - & 1\\
        $1^2\times576$ & conv2d, $1\times1$ & - & 1024 & - & 1\\
        $1^2\times1024$ & conv2d, $1\times1$ & - & k &  - & 1\\
        \hline
    \end{tabular}}
    \caption{Model specification of MicroBotNet (SE indicates if a Squeeze-And-Excite is used, s indicates the stride used).} 
    \label{tab:architecture}
\end{table}

\subsection{Unsupervised Learning Example}
We add a unit norm constraint on $w$ and a KL penalty on $f$ so scores resemble a standard normal. $f$ is not parametrized in these experiments. 
When Dynamic Thresholding is not used, the static noise experiments show FAST and Leaky-SLIPD to be comparable in performance, having many overlaps.
When Dynamic Thresholding is used, for all 3 sequences, Leaky-SLIPD outperforms FAST at 0 noise. For sequence 6, FAST consistently outperforms Leaky-SLIPD on average at all levels of noise except for 0. For sequence 3, Leaky-SLIPD generally outperforms FAST on average for lower noise levels ($\leq 60$ standard deviation), and the FAST marginally outperforms Leaky-SLIPD on average when noise levels exceed 80. On sequence 0, Leaky-SLIPD consistently outperforms FAST on average. Additionally, FAST has significant outliers where it completely fails at certain turns, resulting in extremely high MSE (causing the large standard errors we see in the graph). Overall, Leaky-SLIPD seems at least comparable to FAST, better in some cases and somewhat worse in others. Note: these experiments used slightly different hyperparameters to minimize outlier cases where trajectories were extremely off for each algorithm (typically FAST). The error metric is no longer MSE, but mean euclidean error. In the future, a more expressive, parametrized function (perhaps an MLP) may be useful to consider for improving this method.

\begin{figure}[h]
    \centering
    \includegraphics[width=.95\columnwidth]{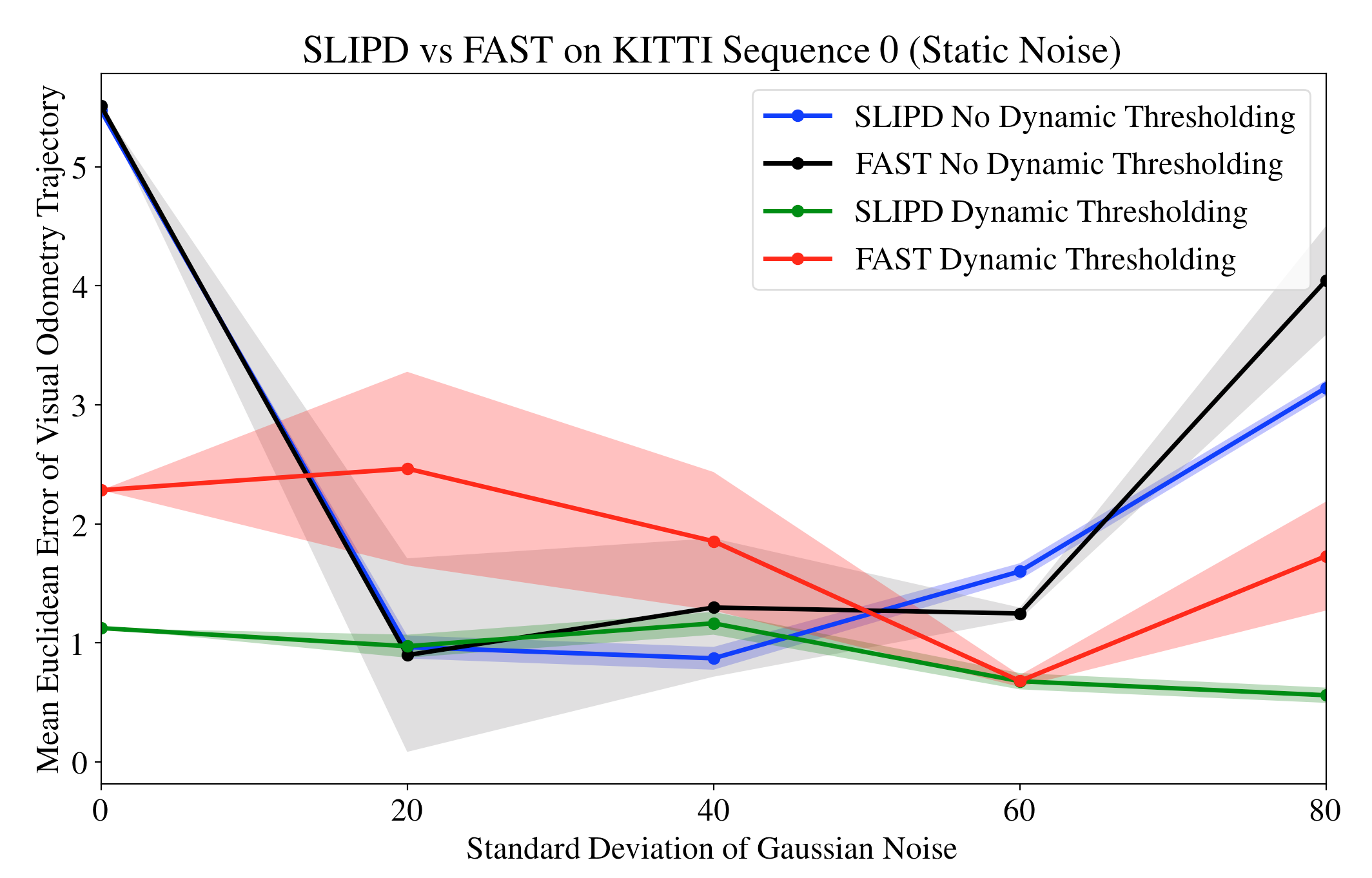}
    \caption{{\small Leaky SLIPD on Sequence 0 with static noise}\\ Without dynamic thresholding, FAST and Leaky-SLIPD perform similarly. With dynamic thresholding, Leaky-SLIPD consistently outperforms FAST (except at $\sigma = 60$) with a significantly lower standard error.}
    \label{fig:static0}
    \vspace{-4mm}
\end{figure}
\begin{figure}[t]
    \centering
    \includegraphics[width=.95\columnwidth]{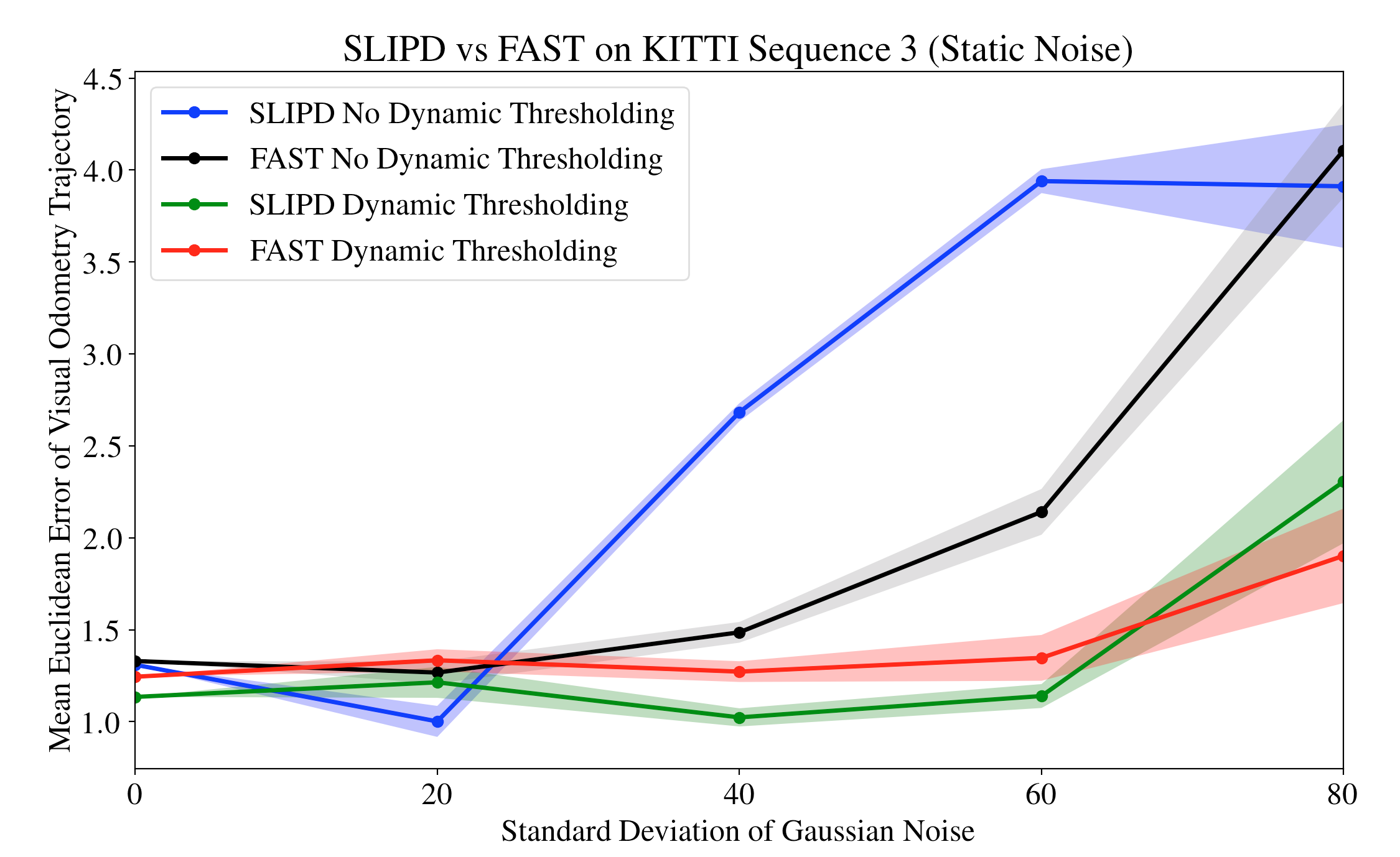}
    \caption{{\small Leaky SLIPD on Sequence 3 with static noise}\\ Without dynamic thresholding, FAST outperforms Leaky-SLIPD. With dynamic thresholding, Leaky-SLIPD consistently outperforms FAST (except at $\sigma = 80$) with a slightly lower standard error overall.}
    \label{fig:static3}
    \vspace{-4mm}
\end{figure}
\begin{figure}[t]
    \centering
    \includegraphics[width=.95\columnwidth]{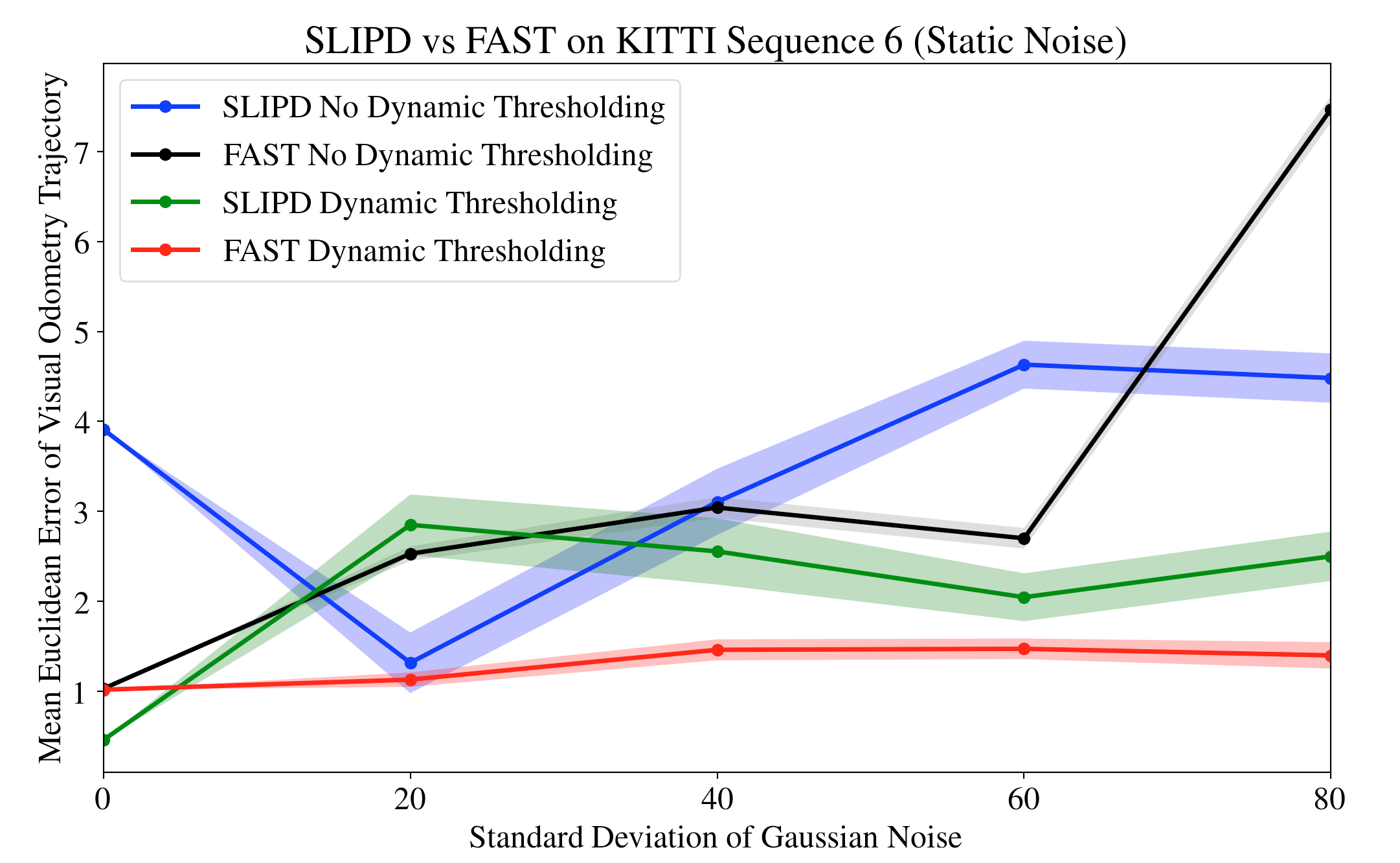}
    \caption{{\small Leaky SLIPD on Sequence 6 with static noise}\\ Without dynamic thresholding, FAST and Leaky-SLIPD perform similarly. With dynamic thresholding, FAST consistently outperforms FAST (except at $\sigma = 0$) with a lower standard error. }
    \label{fig:static6}
    \vspace{-4mm}
\end{figure}

%% file: main.bbl
\begin{thebibliography}{10}
\providecommand{\url}[1]{#1}
\csname url@rmstyle\endcsname
\providecommand{\newblock}{\relax}
\providecommand{\bibinfo}[2]{#2}
\providecommand\BIBentrySTDinterwordspacing{\spaceskip=0pt\relax}
\providecommand\BIBentryALTinterwordstretchfactor{4}
\providecommand\BIBentryALTinterwordspacing{\spaceskip=\fontdimen2\font plus
\BIBentryALTinterwordstretchfactor\fontdimen3\font minus
  \fontdimen4\font\relax}
\providecommand\BIBforeignlanguage[2]{{%
\expandafter\ifx\csname l@#1\endcsname\relax
\typeout{** WARNING: IEEEtran.bst: No hyphenation pattern has been}%
\typeout{** loaded for the language `#1'. Using the pattern for}%
\typeout{** the default language instead.}%
\else
\language=\csname l@#1\endcsname
\fi
#2}}

\bibitem{flynn1987gnat}
A.~M. Flynn, ``Gnat robots (and how they will change robotics),'' in \emph{IEEE
  Micro Robots and Teleoperators Workshop: An investigation of micromechanical
  structures, actuators and sensors, Hyannis, MA, Nov. 9-11, 1987}.\hskip 1em
  plus 0.5em minus 0.4em\relax MIT Artificial Intelligence Laboratory, 1987.

\bibitem{brambilla2013swarm}
M.~Brambilla, E.~Ferrante, M.~Birattari, and M.~Dorigo, ``Swarm robotics: a
  review from the swarm engineering perspective,'' \emph{Swarm Intelligence},
  vol.~7, no.~1, pp. 1--41, 2013.

\bibitem{drew2018toward}
D.~S. Drew, N.~O. Lambert, C.~B. Schindler, and K.~S. Pister, ``Toward
  controlled flight of the ionocraft: a flying microrobot using
  electrohydrodynamic thrust with onboard sensing and no moving parts,''
  \emph{IEEE Robotics and Automation Letters}, vol.~3, no.~4, pp. 2807--2813,
  2018.

\bibitem{contreras2017first}
D.~S. Contreras, D.~S. Drew, and K.~S. Pister, ``First steps of a
  millimeter-scale walking silicon robot,'' in \emph{2017 19th International
  Conference on Solid-State Sensors, Actuators and Microsystems}.\hskip 1em
  plus 0.5em minus 0.4em\relax IEEE, 2017, pp. 910--913.

\bibitem{jafferis2019untethered}
N.~T. Jafferis, E.~F. Helbling, M.~Karpelson, and R.~J. Wood, ``Untethered
  flight of an insect-sized flapping-wing microscale aerial vehicle,''
  \emph{Nature}, vol. 570, no. 7762, p. 491, 2019.

\bibitem{zhang2011bioinspired}
X.~Zhang, J.~Zhao, Q.~Zhu, N.~Chen, M.~Zhang, and Q.~Pan, ``Bioinspired aquatic
  microrobot capable of walking on water surface like a water strider,''
  \emph{ACS applied materials \& interfaces}, vol.~3, no.~7, pp. 2630--2636,
  2011.

\bibitem{zou2016liftoff}
Y.~Zou, W.~Zhang, and Z.~Zhang, ``Liftoff of an electromagnetically driven
  insect-inspired flapping-wing robot,'' \emph{IEEE Transactions on Robotics},
  vol.~32, no.~5, pp. 1285--1289, 2016.

\bibitem{Rentmeister_2020}
J.~S. Rentmeister, K.~Pister, and J.~T. Stauth, ``A 120-330\,{V}, sub-$\mu${A},
  optically powered microrobotic drive {IC} for {DARPA SHRIMP},'' in
  \emph{GOMACTech}, 2020.

\bibitem{scum}
F.~{Maksimovic}, B.~{Wheeler}, D.~C. {Burnett}, O.~{Khan}, S.~{Mesri},
  I.~{Suciu}, L.~{Lee}, A.~{Moreno}, A.~{Sundararajan}, B.~{Zhou}, R.~{Zoll},
  A.~{Ng}, T.~{Chang}, X.~{Villajosana}, T.~{Watteyne}, A.~{Niknejad}, and
  K.~S.~J. {Pister}, ``A crystal-free single-chip micro mote with integrated
  802.15.4 compatible transceiver, sub-mw ble compatible beacon transmitter,
  and cortex m0,'' in \emph{Symposium on VLSI Circuits}, 6 2019, pp. C88--C89.

\bibitem{Choy2003Camera}
J.~Choy, ``A $10\mu \text{J}/$frame $1\text{mm}^2$ $128 \times 128$ cmos active
  image sensor,'' Master's thesis, University of California, Berkeley, 2003.

\bibitem{hanson20100}
S.~Hanson, Z.~Foo, D.~Blaauw, and D.~Sylvester, ``A 0.5 v sub-microwatt cmos
  image sensor with pulse-width modulation read-out,'' \emph{IEEE Journal of
  Solid-State Circuits}, vol.~45, no.~4, pp. 759--767, 2010.

\bibitem{nagabandi2017neural}
A.~Nagabandi, G.~Yang, T.~Asmar, G.~Kahn, S.~Levine, and R.~S. Fearing,
  ``Neural network dynamics models for control of under-actuated legged
  millirobots,'' \emph{Intelligent Robots and Systems}, 2018.

\bibitem{lowlevelmbrl}
N.~O. {Lambert}, D.~S. {Drew}, J.~{Yaconelli}, S.~{Levine}, R.~{Calandra}, and
  K.~S.~J. {Pister}, ``Low-level control of a quadrotor with deep model-based
  reinforcement learning,'' \emph{IEEE Robotics and Automation Letters},
  vol.~4, no.~4, pp. 4224--4230, 2019.

\bibitem{williams2017information}
G.~Williams, N.~Wagener, B.~Goldfain, P.~Drews, J.~M. Rehg, B.~Boots, and E.~A.
  Theodorou, ``Information theoretic mpc for model-based reinforcement
  learning,'' in \emph{2017 IEEE International Conference on Robotics and
  Automation}.\hskip 1em plus 0.5em minus 0.4em\relax IEEE, 2017, pp.
  1714--1721.

\bibitem{janner2019trust}
M.~Janner, J.~Fu, M.~Zhang, and S.~Levine, ``When to trust your model:
  Model-based policy optimization,'' \emph{arXiv preprint arXiv:1906.08253},
  2019.

\bibitem{chua2018deep}
K.~Chua, R.~Calandra, R.~McAllister, and S.~Levine, ``Deep reinforcement
  learning in a handful of trials using probabilistic dynamics models,'' in
  \emph{Advances in Neural Information Processing Systems}, 2018, pp.
  4754--4765.

\bibitem{bailey2006simultaneous}
T.~Bailey and H.~Durrant-Whyte, ``Simultaneous localization and mapping (slam):
  Part ii,'' \emph{IEEE robotics \& automation magazine}, vol.~13, no.~3, pp.
  108--117, 2006.

\bibitem{durrant2006simultaneous}
H.~Durrant-Whyte and T.~Bailey, ``Simultaneous localization and mapping: part
  i,'' \emph{IEEE robotics \& automation magazine}, vol.~13, no.~2, pp.
  99--110, 2006.

\bibitem{rosten2005fast}
E.~{Rosten} and T.~{Drummond}, ``Fusing points and lines for high performance
  tracking,'' in \emph{International Conference on Computer Vision}, vol.~2, 10
  2005, pp. 1508--1515 Vol. 2.

\bibitem{scaramuzza2011visual}
D.~Scaramuzza and F.~Fraundorfer, ``Visual odometry [tutorial],'' \emph{IEEE
  robotics \& automation magazine}, vol.~18, no.~4, pp. 80--92, 2011.

\bibitem{iandola2016squeezenet}
F.~N. Iandola, S.~Han, M.~W. Moskewicz, K.~Ashraf, W.~J. Dally, and K.~Keutzer,
  ``Squeezenet: Alexnet-level accuracy with 50x fewer parameters and 0.5 mb
  model size,'' \emph{arXiv preprint arXiv:1602.07360}, 2016.

\bibitem{howard2017mobilenets}
A.~G. Howard, M.~Zhu, B.~Chen, D.~Kalenichenko, W.~Wang, T.~Weyand,
  M.~Andreetto, and H.~Adam, ``Mobilenets: Efficient convolutional neural
  networks for mobile vision applications,'' \emph{arXiv preprint
  arXiv:1704.04861}, 2017.

\bibitem{qin2018fd}
Z.~Qin, Z.~Zhang, X.~Chen, C.~Wang, and Y.~Peng, ``Fd-mobilenet: improved
  mobilenet with a fast downsampling strategy,'' in \emph{IEEE International
  Conference on Image Processing}.\hskip 1em plus 0.5em minus 0.4em\relax IEEE,
  2018, pp. 1363--1367.

\bibitem{zhu2016trained}
C.~Zhu, S.~Han, H.~Mao, and W.~J. Dally, ``Trained ternary quantization,''
  \emph{arXiv preprint arXiv:1612.01064}, 2016.

\bibitem{howard2019searching}
A.~Howard, M.~Sandler, G.~Chu, L.-C. Chen, B.~Chen, M.~Tan, W.~Wang, Y.~Zhu,
  R.~Pang, V.~Vasudevan, \emph{et~al.}, ``Searching for mobilenetv3,'' in
  \emph{Proceedings of the IEEE International Conference on Computer Vision},
  2019, pp. 1314--1324.

\bibitem{ramachandran2017searching}
P.~Ramachandran, B.~Zoph, and Q.~V. Le, ``Searching for activation functions,''
  \emph{arXiv preprint arXiv:1710.05941}, 2017.

\bibitem{sandler2018mobilenetv2}
M.~Sandler, A.~Howard, M.~Zhu, A.~Zhmoginov, and L.-C. Chen, ``Mobilenetv2:
  Inverted residuals and linear bottlenecks,'' in \emph{Proceedings of the IEEE
  conference on computer vision and pattern recognition}, 2018, pp. 4510--4520.

\bibitem{hu2018squeeze}
J.~Hu, L.~Shen, and G.~Sun, ``Squeeze-and-excitation networks,'' in
  \emph{Proceedings of the IEEE conference on computer vision and pattern
  recognition}, 2018, pp. 7132--7141.

\bibitem{horowitzenergy}
M.~Horowitz, ``Energy table for 45nm process,'' Stanford VLSI wiki.

\bibitem{bankman2018always}
D.~Bankman, L.~Yang, B.~Moons, M.~Verhelst, and B.~Murmann, ``An always-on 3.8
  uj 86\% cifar-10 mixed-signal binary cnn processor with all memory on chip in
  28-nm cmos,'' \emph{IEEE Journal of Solid-State Circuits}, vol.~54, no.~1,
  pp. 158--172, 2018.

\bibitem{yang2018learning}
B.~Yang, G.~Wang, R.~Calandra, D.~Contreras, S.~Levine, and K.~Pister,
  ``Learning flexible and reusable locomotion primitives for a microrobot,''
  \emph{IEEE Robotics and Automation Letters}, vol.~3, no.~3, pp. 1904--1911,
  2018.

\bibitem{mismatch2019}
N.~O. Lambert, B.~D. Amos, O.~Yadan, and R.~Calandra, ``Objective mismatch in
  model-based reinforcement learning,'' \emph{arXiv preprint arXiv:}, 2019.

\bibitem{bitcraze2016crazyflie}
A.~Bitcraze, ``Crazyflie 2.0,'' 2016.

\bibitem{Geiger2012CVPR}
A.~Geiger, P.~Lenz, and R.~Urtasun, ``Are we ready for autonomous driving? the
  kitti vision benchmark suite,'' in \emph{Conference on Computer Vision and
  Pattern Recognition}, 2012.

\bibitem{quadnet}
\BIBentryALTinterwordspacing
N.~Savinov, A.~Seki, L.~Ladicky, T.~Sattler, and M.~Pollefeys, ``Quad-networks:
  unsupervised learning to rank for interest point detection,'' \emph{CoRR},
  vol. abs/1611.07571, 2016. [Online]. Available:
  \url{http://arxiv.org/abs/1611.07571}
\BIBentrySTDinterwordspacing

\bibitem{krizhevsky2014cifar}
A.~Krizhevsky, V.~Nair, and G.~Hinton, ``The cifar-10 dataset,'' \emph{online:
  http://www. cs. toronto. edu/kriz/cifar. html}, vol.~55, 2014.

\bibitem{thop}
``Thop: Pytorch-opcounter,''
  \url{https://github.com/Lyken17/pytorch-OpCounter/}, accessed: 2020-02-26.

\bibitem{ostfeld2016high}
A.~E. Ostfeld, A.~M. Gaikwad, Y.~Khan, and A.~C. Arias, ``High-performance
  flexible energy storage and harvesting system for wearable electronics,''
  \emph{Scientific reports}, vol.~6, p. 26122, 2016.

\end{thebibliography}
